\documentclass[10pt, a4paper]{article}

\usepackage{lrec2026}
\usepackage{booktabs}
\usepackage{multirow}
\usepackage{xurl}
\usepackage{xcolor}
\usepackage{tabularx}
\usepackage{array}
\usepackage{rotating}

\title{Open but Incompatible: A License Compatibility Analysis of Corpora for Low-Resource African Languages}

\name{Ernst van Gassen}
\address{Arktos AI Lab \\
         Netherlands \\
         evg.gassen@gmail.com}

\abstract{
Creative Commons licenses dominate African NLP corpus releases, but their compatibility rules are rarely applied. CC-BY-SA and CC-BY-NC cannot be combined in a single published dataset; a NoDerivs clause silently prohibits tokenisation and annotation. This paper audits the license provenance of over twenty corpus families used in African NLP, constructs a six-tier compatibility matrix, and applies it to three case-study languages: Kituba/Munukutuba, Zarma, and Moore. Four failure modes are documented with primary-source evidence: outright prohibition (JW300, removed from OPUS after a legal audit confirmed Terms of Service violation); composite license misrepresentation (WAXAL, whose CC-BY 4.0 claim is contradicted by its own HuggingFace dataset card); a NoDerivs clause hidden behind a CC-BY label (Tanzil); and data persistence failure (the Congolese Radio Corpus, where 402 of 405 source URLs are now dead). A pre-annotation due diligence checklist and a survey of legally clean enrichment opportunities close the paper.
\\ \newline  }

\begin{document}

\maketitleabstract

\section{Introduction}

NLP researchers are not lawyers. For high-resource languages, this is ordinarily not a problem. Corpora in common use have been legally vetted over decades of practice. For low-resource African languages, neither condition holds.

Since 2019, parallel corpora, NER datasets, sentiment benchmarks, and speech resources have been published for dozens of African languages. This paper argues that much of this output has not received systematic license review. The consequences are beginning to surface. JW300 \citep{Agic2019JW300} was a parallel corpus covering 300+ languages, including dozens with no alternative source. It was found to have been built in violation of the Jehovah's Witnesses website Terms of Service, which explicitly prohibit text and data mining. A legal audit by the Centre for Intellectual Property and Information Technology (CIPIT) in Nairobi confirmed the violation \citep{CIPIT2020}. OPUS removed the corpus. Every dataset, model, and benchmark that incorporated JW300 data now carries a contaminated provenance chain.

JW300 is not an isolated case. GEITje, a Dutch-language model, was removed from HuggingFace following a copyright enforcement request from Stichting BREIN \citep{Rijgersberg2023GEITje,NuNL2024BREIN,RTLNL2024GEITje,Tweakers2024BREIN,GoingDutch2024GEITje}; for high-resource languages these incidents are disruptive but recoverable. For African languages, losing one corpus may mean losing the only usable source for that language. Common Corpus \citep{Langlais2025CommonCorpus}, a two-trillion-token corpus explicitly curated for open licensing, illustrates the baseline: an audit of its training table finds 15 native Sub-Saharan languages accounting for approximately 91 rows combined, against 18,485 for English (Section~\ref{sec:common_corpus_audit}).

The stakes are higher for low-resource languages than for other NLP contexts for reasons that compound: there is no substitution corpus when a source is lost; annotation investment is sunk; one license conflict can block half the available landscape; and tainted sources cascade through multi-language benchmark releases.

Three languages anchor the case studies. \textbf{Kituba/Munukutuba} (\texttt{ktu}/\texttt{mkw}), the vehicular language of southern Congo-Brazzaville and a Kinshasa lingua franca, has 5--8 million speakers and no entry in FLORES-200, MasakhaNER, or AfriSenti. \textbf{Zarma} (\texttt{dje}), spoken by 4--5 million people in southwestern Niger, has MT and NER resources from the 27Group but is absent from FLORES-200, NLLB, and all major benchmarks; OCHA and UNHCR list it as a required crisis communication language. \textbf{Moore} (\texttt{mos}), with FLORES-200 coverage and active research groups in Burkina Faso, serves as contrast: the upper bound of what the open-license landscape currently offers a relatively well-studied but still low-resource language.

\section{Related Work}

\paragraph{Legal scholarship on open licensing.}
Creative Commons licensing has attracted sustained scholarly critique. \citet{Katz2006Pitfalls} identifies two structural problems: variant proliferation creates user confusion, and ShareAlike terms create compatibility deadlocks that prevent legal distribution of derivatives. \citet{Boyle2003Enclosure} provides the theoretical frame: restrictive IP licensing constitutes a second enclosure movement.

\paragraph{Data provenance and license audits in NLP.}
\citet{Gebru2021Datasheets} and \citet{Bender2021DataStatementsNLP} propose structured documentation standards for datasets and NLP corpora. Both argue that provenance and licensing must be documented as first-class metadata. \citet{Dodge2021Documenting} applies this to large webtext corpora, finding machine-generated text and benchmark contamination that documentation would have surfaced. \citet{Kreutzer2022Quality} audits 205 language-specific corpora across five multilingual web-crawl datasets, finding at least 15 contain no usable text and many use ambiguous language codes. The challenges of web-mined corpora for LLM pre-training are surveyed in \citet{Perelkiewicz2024WebMined}.

The closest prior work to this paper is the Data Provenance Initiative \citep{Longpre2023Provenance,Longpre2024Nature}. It audits 1,800+ text datasets used to train LLMs. License omission rates are above 70\%; error rates are above 50\%. The Data Provenance Initiative operates at the general-purpose LLM level. It does not focus on African languages. It does not construct a license compatibility matrix. It does not address the failure modes documented here. \citet{Mahari2024Discit} extends this line of work into legal analysis, arguing that provenance documentation affects fair use claims for fine-tuning data.

\paragraph{Legal analysis of AI training data.}
\citet{Henderson2023FairUse} analyses the four US fair use factors as applied to foundation model training, concluding that fair use is plausible but not guaranteed. \citet{Lee2023Talkin} maps copyright questions across the full generative-AI supply chain. \citet{Jernite2022Governance} proposes a multi-stakeholder data governance framework addressing licensing at each stage of the data journey. None of these works applies this framework to low-resource African languages.

\paragraph{African NLP data licensing.}
\citet{Nekoto2020Participatory} is the founding Masakhane paper. It is the first African NLP work to address data ownership and licensing governance explicitly. It brought the JW300 licensing issue to community attention and motivated the subsequent CIPIT legal audit \citep{CIPIT2020}. \citet{Adelani2021MasakhaNER,Adelani2022MasakhaNER2} document licensing decisions for MasakhaNER releases. \citet{OkerieMarivate2024} surveys the African NLP community on copyright barriers. It finds that the JW300 withdrawal created downstream disruption for projects with no alternative sources. \citet{Omino2025NOODL} proposes the Nwulite Obodo Open Data License (NOODL), a tiered community license designed for African language datasets.

\citet{Tiedemann2020Tatoeba} demonstrates the value of Tatoeba for low-resource MT benchmarking; this paper focuses on the African-language subset and the licensing constraints governing which sources can be legally combined. The compatibility matrix in Section~\ref{sec:compat} is the practical output of that synthesis.

\section{License Taxonomy}

I define six license tiers for African NLP text corpora, ordered from least to most restrictive. For non-specialist readers: \textbf{NC} (Non-Commercial) means the resource may not be used for revenue-generating purposes as defined by the license; what constitutes commercial use is context-dependent and jurisdiction-sensitive, but publishing an annotated dataset via a paid service or commercial API is a clear case. \textbf{ND} (NoDerivs) means the license prohibits \emph{sharing} modified, adapted, tokenised, or otherwise derived versions; private use may still be permitted by applicable law (e.g.\ fair use, TDM exceptions), but no published annotation dataset derived from an ND source can be legally distributed under the license terms.

\begin{table}[!ht]
\centering
\small
\begin{tabularx}{\columnwidth}{|l|X|c|}
\hline
\textbf{Tier} & \textbf{Description} & \textbf{Risk} \\
\hline
T1 & CC0 / public domain / government text. No restrictions. & None \\
\hline
T2 & CC-BY / MIT / Apache 2.0. Attribution required; no share-alike, no NC, no ND. & Low \\
\hline
T3 & CC-BY-SA. Share-alike propagates to all published derivatives. & Medium \\
\hline
T4a & CC-BY-NC. Non-commercial restriction (NC): the resource may not be used for revenue-generating purposes. Incompatible with T3. & High \\
\hline
T4b & CC-BY-ND or CC-BY with undisclosed ND clause. NoDerivs (ND): the license prohibits distributing modified or adapted versions; publishing an annotated dataset derived from a T4b source is not permitted under the license, regardless of the output license chosen. & High \\
\hline
T5 & ToS violation, permission denied, or copyright holder prohibition. & \textbf{Prohibited} \\
\hline
\end{tabularx}
\caption{License tier taxonomy. T4a and T4b are separated because their restrictions operate differently and are mutually incompatible with T3.}
\label{tab:tiers}
\end{table}

The key practical distinction is between T3 (share-alike propagates but derivatives are permitted) and T4b (derivatives not permitted at all). Many practitioners conflate these, treating all non-T2 sources as merely requiring a more restrictive output license. This is incorrect: a T4b source cannot legally be incorporated into any published annotation dataset, regardless of the output license chosen.

\section{License Compatibility Matrix}
\label{sec:compat}

Table~\ref{tab:compat} shows the legally valid output license when two corpus sources are combined. ``$\times$'' denotes an incompatible combination: no single license can satisfy both sources' requirements simultaneously.

\begin{table}[!ht]
\centering
\small
\begin{tabular}{l|cccccc}
\hline
 & \textbf{T1} & \textbf{T2} & \textbf{T3} & \textbf{T4a} & \textbf{T4b} & \textbf{T5} \\
\hline
\textbf{T1} & T1+ & T2 & T3 & T4a & $\times$ & $\times$ \\
\textbf{T2} & T2  & T2 & T3 & T4a & $\times$ & $\times$ \\
\textbf{T3} & T3  & T3 & T3 & $\times$ & $\times$ & $\times$ \\
\textbf{T4a}& T4a & T4a& $\times$ & T4a & $\times$ & $\times$ \\
\textbf{T4b}& $\times$ & $\times$ & $\times$ & $\times$ & $\times$ & $\times$ \\
\textbf{T5} & $\times$ & $\times$ & $\times$ & $\times$ & $\times$ & $\times$ \\
\hline
\end{tabular}
\caption{License compatibility matrix. Cell shows the required output license when the row-source and column-source are combined. $\times$ = incompatible combination; no valid output license exists. T1+ = any license acceptable. T4b and T5 are incompatible with all other tiers.}
\label{tab:compat}
\end{table}

A note on provenance quality independent of license tier: the compatibility matrix captures output license requirements, not the trustworthiness of the collection process. Two datasets can both carry CC0 labels while having very different provenance standing. ParaCrawl \citep{Banon2020ParaCrawl} is a web-scale parallel corpus co-financed by the EU Connecting Europe Facility, with the University of Edinburgh as lead institution, and carries CC0. Importantly, ParaCrawl explicitly states that it does not own the underlying text; CC0 applies to the packaging and database rights only. Its institutional context provides a degree of accountability that informal web scrapes do not. JW300 also presented as open-access but was built in violation of platform Terms of Service. The license tier alone does not distinguish these two cases; provenance standing does.

\subsection{Dataset License versus Redistribution Rights}
\label{sec:redistribution}

Several widely-used corpora are web-mined, meaning the dataset license reflects packaging or database rights rather than rights in the underlying text \citep{Perelkiewicz2024WebMined}. CCMatrix \citep{Schwenk2021CCMatrix} is mined from Common Crawl snapshots and carries no stated text license on OPUS or in its paper. NLLB mined bitext \citep{NLLB2022} uses Common Crawl WET files as a primary source; ODC-BY, the license on NLLB bitext, governs database rights, not rights in the underlying text. WURA \citep{Oladipo2023WURA} is built by auditing mC4 (itself Common Crawl-derived) plus additional focused crawls. For these corpora, the dataset-level license does not clear the underlying text for redistribution or relicensing.

A use-case distinction is practically consequential. Model \textit{training} on mC4-derived text may be defensible under fair use or EU Text and Data Mining (TDM) exceptions, depending on jurisdiction. Publishing an \textit{annotated dataset} derived from the same text is redistribution of copyrighted content. The dataset's Apache 2.0 or CC packaging label does not change this: it covers the packaging, not the text. This distinction is not visible in the compatibility matrix. For the primary output of the African NLP community (published annotation datasets), redistribution risk applies to all mC4-derived sources regardless of their stated license. Rights-cleared sources (UDHR, TICO-19, FLEURS, SMOL, original speech recordings) avoid this risk entirely. Practitioners who use WURA or Leipzig as annotation seeds for a published NER or POS dataset are redistributing copyrighted web text under a license the copyright holders never granted.

One useful indicator of provenance quality is institutional context. EU-funded projects with named university leads, corpora with public ethics disclosures, and datasets with named investigators tend to carry more accountability than anonymous uploads. This is an indicator of lower risk, not a guarantee. No institutional label substitutes for verification of the actual collection method.

For African languages, ParaCrawl's bonus releases include English--Swahili (132,517 sentence pairs, CC0) and English--Somali (14,879 sentence pairs, CC0).

Three results from this matrix are practically consequential for African NLP:

\textbf{(1) T3 $\times$ T4a = incompatible.} Wikipedia (CC-BY-SA, T3) and the 27Group Feriji Zarma corpus (CC-BY-NC, T4a) cannot be combined in a published dataset. A practitioner who annotates Wikipedia sentences alongside Feriji sentences and publishes the result has created a legally invalid dataset. Wikipedia's share-alike requirement demands CC-BY-SA output. Feriji's non-commercial restriction demands CC-BY-NC output. No single license satisfies both.

\textbf{(2) T4b $\times$ anything = blocked.} Any corpus with a NoDerivs clause, including Tanzil, cannot legally be used to create a published annotation dataset. The annotation itself is a derivative work. This is not a question of the output license; distributing a derived annotation dataset is not permitted under the license, regardless of jurisdiction-specific exceptions that may apply to private use.

\textbf{(3) T3 propagates upward; T4a similarly.} A T2 source combined with a T3 source produces a T3 output. A T2 or T3 source combined with a T4a source produces a T4a output (non-commercial restriction is inherited). Practitioners who use Wikipedia as annotation seed must release under CC-BY-SA 4.0. MasakhaNER 2.0 \citep{Adelani2022MasakhaNER2} uses Wikipedia text in its annotation pipeline. Note that MasakhaNER's HuggingFace dataset card lists CC BY-NC 4.0 for the dataset release; the source-text licensing is heterogeneous. Practitioners should verify the specific version they use. The important point is that license decisions must be made \textit{before} annotation begins. Choosing CC-BY-SA forecloses certain commercial downstream uses.

\section{African NLP Corpus Survey}
\label{sec:survey}
\label{sec:audit}

Table~\ref{tab:corpus_survey} (Appendix~\ref{app:corpus_survey}) surveys the corpus families used in African NLP with their tier assignments; an asterisk (*) marks web-mined corpora where the dataset license covers packaging or database rights, not rights in the underlying text.

\subsection{Common Corpus: African Language Representation}
\label{sec:common_corpus_audit}

We streamed the full Common Corpus training split \citep{Langlais2025CommonCorpus} and filtered by the \texttt{language} field. All 91 native Sub-Saharan rows carry CC-BY-SA licenses; \texttt{subset} and \texttt{url} fields are \texttt{null} throughout. Text content identifies the source as Wikipedia 2023 via MediaWiki markup (e.g.\ \texttt{\{\{infobox tan\`{a}na\}\}} in Malagasy rows). The 16 rows labelled ``Various open data'' (Lingala 7, Kabyle 5, Wolof 3, Hausa 1) are language-identification errors on French archival documents; none contains usable African-language text. Common Corpus is not an independent African-language source: it repackages the same Wikipedia dumps audited in Table~\ref{tab:corpus_survey}, with worse provenance metadata. Researchers should not count both as separate entries. The ratio is approximately 200:1 (English 18,485 rows; all 15 native Sub-Saharan languages combined, 91 rows); Afrikaans alone, with 11 rows, exceeds the native total. This figure also reflects a structural feedback loop: platforms such as YouTube generate CC-licensed ASR transcripts only for languages with an existing seed model. Swahili is the sole Sub-Saharan language with YouTube ASR support; it accumulates more CC text with each upload while the flywheel never starts for Lingala, Kikongo, or Tshiluba. Critical mass of labelled speech data is a prerequisite, not merely a goal.

\subsection{Applying the Compatibility Matrix to Case-Study Languages}

Table~\ref{tab:compatibility_cases} shows which source combinations are legally valid for the three case-study languages, and what output license each combination requires.

\begin{table}[!ht]
\centering
\small
\begin{tabularx}{\columnwidth}{Xll}
\hline
\textbf{Combination} & \textbf{Valid?} & \textbf{Output license} \\
\hline
\multicolumn{3}{l}{\textit{Kituba (ktu/mkw)}} \\
Leipzig mkw + SMOL ktu & Yes & T2 (CC-BY) \\
Leipzig mkw + kgwiki (CC-BY-SA) & Yes & T3 (CC-BY-SA) \\
Leipzig mkw + Mozilla TTS mkw (NOODL) & \textbf{No} & T4b: no derivatives \\
\hline
\multicolumn{3}{l}{\textit{Zarma (dje)}} \\
MT560 dje + 27Group NER & Yes & T2 (CC-BY 4.0) \\
MT560 dje + 27Group noisy GEC & Yes & T3 (CC-BY-SA 4.0) \\
MT560 dje + 27Group Feriji & Yes & T4a (CC-BY-NC 4.0) \\
Wikipedia (0) + Feriji & N/A & No Wikipedia edition \\
Feriji + 27Group GEC (T3) & \textbf{No} & T4a $\times$ T3 = incompatible \\
\hline
\multicolumn{3}{l}{\textit{Moore (mos)}} \\
MT560 mos + FLORES-200 mos & Yes & T3 (CC-BY-SA 4.0) \\
MT560 mos + MooreFRCollections & Yes & T2 (CC-BY 4.0) \\
MT560 mos + MooresentimentAuto & Yes & T2 (MIT) \\
FLORES-200 + WURA mos & Yes & T3 (CC-BY-SA 4.0) \\
\hline
\end{tabularx}
\caption{Compatibility analysis for case-study language combinations. ``No'' entries indicate legally invalid combinations.}
\label{tab:compatibility_cases}
\end{table}

For Moore, the license landscape is relatively clean. The main sources (MT560, FLORES-200, WURA, MooreFRCollections) are all T2 or T3 and compatible. One caveat applies: the 125,695-row Moore sentiment dataset (\texttt{michsethowusu/mossi-sentiments-corpus}) assigns labels via English back-translation through DistilBERT. In the NLP annotation literature, \textit{silver} labels are automatically generated, while \textit{gold} labels are human-annotated directly in the target language. This dataset should not be used as a gold standard for sentiment benchmarking without human verification of a stratified sample.

For Zarma, the combination of Feriji (T4a) with the 27Group noisy GEC corpus (T3) is incompatible. Both corpora are published by the same research group. A practitioner who used both in a single annotation pipeline would have created a dataset with no valid output license.

\section{Four Failure Modes}
\label{sec:failures}

\subsection{Prohibition: JW300}

JW300 \citep{Agic2019JW300} was a parallel corpus covering 300+ languages, built from the Jehovah's Witnesses website \texttt{jw.org}. It was widely used in African NLP from 2019 onward due to coverage of languages with no other parallel text.

The legal problem is straightforward. The \texttt{jw.org} Terms of Service explicitly prohibit text and data mining. A legal audit by CIPIT Nairobi confirmed the violation \citep{CIPIT2020}. OPUS removed the corpus following Masakhane's formal request for permission, which the organisation denied \citep{WalledCulture2020}. This is Tier 5 under the taxonomy above: prohibited regardless of how the corpus was obtained.

The consequential risk for current practitioners is indirect. Language models, cross-lingual embeddings, and benchmark systems trained before 2021 may incorporate JW300-derived representations. Derivative datasets built from those models carry a tainted provenance chain. African NLP papers should include an explicit statement: \textit{``This dataset does not include JW300-derived text or derivatives thereof.''} This is analogous to the ``IRB approved'' statement in human subjects research: a reproducible declaration that reviewers can verify.

\subsection{Composite License Misrepresentation: WAXAL}

WAXAL \citep{Diack2026WAXAL} is a 2026 speech dataset covering 19 African languages for ASR and 16 for TTS. The arXiv paper \citep{Diack2026WAXAL} claims that the collection is released under a uniform CC-BY 4.0 license. The HuggingFace dataset card \citep{WAXALDatasetCard} explicitly contradicts this, listing per-provider licenses. Per-provider attribution, traced from WAXAL's supplementary tables:

\begin{itemize}
  \item \textbf{CC-BY 4.0 (T2):} University of Ghana contributions only: Akan, Ewe, Dagbani, Dagaare, Ikposo (ASR); Fante, Twi (TTS).
  \item \textbf{CC-BY-SA 4.0 (T3):} All other contributions: Makerere University (Acholi, Luganda, Masaaba, Nyankole, Soga), Digital Umuganda (Fula, \textbf{Lingala}, Shona, Malagasy, Amharic, Oromo, Sidama, Tigrinya, Wolaytta), Media Trust (Fula, Igbo, Hausa, Yoruba, Nigerian Pidgin), Loud and Clear (Kikuyu, Luo).
\end{itemize}

The composite misrepresentation creates a concrete legal failure. A practitioner who reads the WAXAL arXiv abstract, downloads the Lingala subset, annotates a NER dataset from its transcripts, and publishes under CC-BY 4.0 has violated the CC-BY-SA 4.0 share-alike requirement of the Digital Umuganda contribution. This occurs despite acting in good faith on the stated license. Lingala, Hausa, Igbo, Yoruba, and all Makerere-sourced languages in WAXAL require CC-BY-SA 4.0 output for any published derivative.

A noteworthy asymmetry exists. Digital Umuganda's standalone \textsc{AfriVoice} dataset, which covers the same Lingala recordings, is released under CC-BY 4.0. The same speech data carries different terms depending on which dataset packaging it is accessed through. Practitioners cannot resolve this without per-provider provenance tracing that the arXiv paper does not facilitate.

Composite dataset papers should include a per-language provenance and license table as a required metadata artifact.

\subsection{Hidden NoDerivs Restriction: Tanzil}

Tanzil (\texttt{tanzil.net}) provides Quran translations in approximately 40 languages, including several African languages (Hausa, Swahili, Somali, Amharic, partial Yoruba). Its stated license is CC-BY 3.0. In the NLP literature, CC-BY is typically treated as Tier 2: permissive, derivatives allowed, attribution required.

The Tanzil license page explicitly states: \textit{``You are not allowed to modify this text in any way''} \citep{TanzilLicense}. This is a NoDerivs restriction (Tier 4b). It is not disclosed in the CC-BY label. A practitioner who tokenises Tanzil text, aligns it to a parallel target, and publishes the result as a training dataset has violated this restriction. The derivative prohibition applies regardless of the output license chosen.

The NoDerivs clause reflects the religious status of the Quran in Islam. Tanzil's policy holds that Quranic text may not be altered, in order to preserve the integrity of a text considered holy and inviolable in Islamic tradition \citep{TanzilLicense}. This explains why the ND restriction is applied even under an otherwise permissive CC-BY label.

An annotation dataset derived from Tanzil text cannot legally be published under any open license. The NoDerivs clause prohibits the derivative work entirely. To the extent that Tanzil-derived text has been incorporated into African NLP pipelines for languages with Quran translation coverage, those pipelines carry this undisclosed legal risk.

No modern African-language Quran translation is available in a clearly public-domain or CC-BY (without ND) machine-readable format. The classical English translations of Sale (1734, Project Gutenberg \#7440), Rodwell (1861, \#3434), and Palmer (1880, Wikisource) are public domain. These English public domain translations provide no African-language text and are of no direct utility for practitioners building African-language NLP resources.

\subsection{Data Persistence Failure: The Congolese Radio Corpus}

The Congolese Radio Corpus (CRC) \citelanguageresource{CRC} for Lingala was published with a claim of hundreds of hours of broadcast audio sourced from YouTube. An audit conducted in February 2026 found that \textbf{402 of 405 YouTube IDs referenced in the CRC are now dead}, returning 404 errors due to video removal or channel deletion. The reproducible resource is approximately 14.4 hours of elicited LRSC speech and Radio Okapi broadcasts.

This is not a criticism of the original authors. It is a structural warning: \textbf{corpora that depend on third-party platform URLs are inherently non-persistent.} A published corpus that cannot be reproduced by a subsequent researcher is not a scientific contribution in the standard sense. The CRC is not an isolated case. Common Voice drops recordings when contributors withdraw consent. HuggingFace datasets are occasionally removed by their owners. YouTube channels are deleted routinely.

Corpora distributed via repositories with persistent identifiers, such as Zenodo DOIs, OpenSLR stable IDs, or LDC catalogue numbers, have remained reproducible across years. I recommend that African NLP publication venues adopt a data availability standard requiring either (a) a persistent DOI-backed deposit for all corpus resources, or (b) an explicit statement of which components are platform-dependent and may become unavailable.

A related failure is the lack of provenance documentation in community HuggingFace uploads. Several large parallel corpora for African languages carry labels such as ``MT560/OPUS-derived'' with no source URLs, translation pipeline documentation, or quality filter parameters. For example: \texttt{michsethowusu/english-luba-kasai\_sentence-pairs\_mt560} (292,000 rows, CC-BY 4.0), \texttt{michsethowusu/english-congo-swahili\_sentence-pairs\_mt560} (272,000 rows, CC-BY 4.0), and \texttt{michsethowusu/english-zarma\_sentence-pairs\_mt560} (60,000 rows, CC-BY 4.0) all fall into this category. These cannot be audited for license provenance. A practitioner cannot verify whether T5 sources were included in the pipeline, making them legally ambiguous despite carrying permissive license labels.

\paragraph{Data persistence and digital sovereignty.}
The CRC failure is not merely a technical problem; it is also a sovereignty problem. All four platforms implicated in African NLP data rot, YouTube, HuggingFace, GitHub, and OPUS, are maintained by US or European organisations with no specific accountability to African language communities. When a corpus disappears from these platforms, no African institution has the mandate or authority to recover it. This argues for African-controlled digital infrastructure for language data. Initiatives such as SADILAR (South African Centre for Digital Language Resources) and the ISLRN persistent identifier system point toward a model where African language resources are deposited in African-managed archives with ISLRN identifiers \citep{Nekoto2020Participatory,Omino2025NOODL}. The CRC case illustrates what happens in the absence of such infrastructure: a published corpus becomes an unverifiable claim.

\section{Enrichment Opportunities Within the Open-License Landscape}
\label{sec:enrichment}

The foregoing analysis could be read pessimistically. The legal constraints are severe, several documented corpora contain license problems, and authentic open-license text for under-resourced African languages is limited. The opposite reading is more productive: identifying the legal constraints clearly is what makes enrichment tractable.

\paragraph{Transcribing untranscribed speech.}
WAXAL includes speech subsets for which transcripts are not released. The University of Ghana subsets (Akan, Ewe, Dagbani, Dagaare, Ikposo ASR; Fante, Twi TTS) carry CC-BY 4.0 licensing. Transcribing these recordings with community annotators and releasing under CC-BY 4.0 would produce genuinely new, derivative-safe text corpora requiring no new data collection.

\paragraph{Annotation of existing T2/T3 seeds.}
For each case-study language, T2 seed text exists and can be annotated for NER, POS, or sentiment. The key legal decision is whether to include Wikipedia (T3, forcing CC-BY-SA 4.0 output) or restrict annotation to T2 sources (permitting CC-BY 4.0 output). This decision must be made before annotation begins, as it affects downstream commercial usability. A further distinction applies to web-mined T2 sources such as WURA and Leipzig: their packaging license does not clear the underlying text for redistribution. Publishing an annotated dataset whose seed sentences come from WURA is redistribution of mC4-derived content. Rights-cleared T2 sources (FLEURS, SMOL, TICO-19) do not carry this risk and are preferable as annotation seeds when they provide sufficient coverage.

For Kituba, the Leipzig Corpora Collection \citep{Goldhahn2012Leipzig} \texttt{mkw\_community\_2017} entry (143,476 sentences, CC-BY, T2) is, to the author's knowledge, the largest available Kituba text corpus and has not appeared in published NLP work. Combined with the SMOL \texttt{gatitos\_\_en\_ktu} pairs (863 sentences, CC-BY 4.0, T2), it provides an NER annotation seed with known provenance that permits CC-BY 4.0 output without share-alike propagation.

For Zarma, the MT560 parallel corpus (60,515 sentences, CC-BY 4.0, T2) is the safe annotation seed for CC-BY 4.0 output. The 27Group noisy GEC corpus (508,869 sentences, T3) is available but forces CC-BY-SA 4.0 output and is incompatible with Feriji (T4a).

\paragraph{Parallel and bridged resources.}
For zero-pivot African--African pairs, the UDHR (T1, public domain) provides the same 30 articles across 570+ language editions in sentence-aligned form on OPUS; any two editions pair directly with no English or French intermediary. NTREX-128 \citep{Federmann2022NTREX} provides 1,997 professionally translated news sentences in 24 African languages under CC-BY-SA 4.0; the shared source enables direct pairing of any two. For languages outside these resources, bridge construction via FLORES-200 (T3, CC-BY-SA) or TICO-19 (T1, CC0) is possible where both languages have segments against the same pivot. English, French, Arabic, and Portuguese cover the main regional pivot groups. Global Voices (OPUS, CC-BY 3.0) provides human-translated Swahili ($\approx$20K pairs) and Amharic ($\approx$1K). None substitutes for large training corpora, but all are legally clean and available today.

\paragraph{The kgwiki discovery.}
A finding with direct enrichment implications: the Kongo Wikipedia (\texttt{kgwiki}), labeled and indexed as \textit{Kongo}, is written in Kituba/Munukutuba. This is confirmed by article content inspection and the \texttt{Svngoku} dataset card, which explicitly invites speakers of Munukutuba, Kituba, and Kikongo ya Leta to contribute. As a result, 1,200+ articles of usable Kituba text (CC-BY-SA 4.0, T3) may have been missed by practitioners searching for Kituba text under its correct ISO codes (\texttt{ktu}, \texttt{mkw}). The same mislabeling affects FLORES-200's \texttt{kon\_Latn} entry. Researchers building Kongo NLP systems may have trained on Kituba data; researchers building Kituba systems may have missed this resource. Resolving this ISO code confusion, which involves at least five codes (\texttt{kon}/\texttt{kg}, \texttt{ktu}, \texttt{mkw}, \texttt{kwy}), is a prerequisite for any systematic enrichment effort.

\section{A Legal Due Diligence Checklist}

Four steps before annotation begins:

\textbf{Step 1: Inventory sources.} Consult: Wikipedia, Leipzig, UDHR, Tatoeba, FLORES-200, FLEURS, WAXAL (per-provider), WURA, eBible.org (per-translation), African Storybook (per-story), TICO-19, Common Voice (per-subset), MT560/HuggingFace (with provenance caveat), OPUS (excluding JW300-derived subsets). Avoid or flag: CCMatrix (no stated license), TED2020 (T4b), JW300 (T5), bible-uedin (CC0 claimed; verify per-translation rights).

\textbf{Step 2: Assign tiers; distinguish training from redistribution.} Use Table~\ref{tab:tiers}. Do not trust aggregator license labels: Tanzil is T4b despite stating CC-BY; WAXAL majority subsets are T3 despite stating CC-BY 4.0. For web-mined sources (WURA, Leipzig, ParaCrawl, NLLB bitext), distinguish model training (may be defensible under fair use or TDM exceptions) from dataset publication (redistribution, higher risk). Verify against the original source repository.

\textbf{Step 3: Run the compatibility matrix.} Use Table~\ref{tab:compat} for each pair of sources. Any $\times$ must be resolved before annotation begins---usually by dropping one source.

\textbf{Step 4: Verify ISO codes and archive.} For Bantu languages, inspect actual text samples and cross-reference with Ethnologue and Glottolog; kgwiki is Kituba, not Kongo. Record dataset name, license version, retrieval date, and checksum or DOI for every source. Deposit a snapshot in a persistent archive. Include an explicit statement that no JW300-derived text is present.

\section{Discussion}

None of the errors documented here was wilful; each was a legal assumption that NLP practice gave no reason to question explicitly. The compatibility matrix (Table~\ref{tab:compat}) requires no legal expertise: it is a lookup table. Tier assignments require a one-time provenance check per corpus. The checklist requires discipline. A common concern is that Wikipedia's CC-BY-SA license propagates share-alike and is therefore too restrictive. In practice, Wikipedia has 3--5 named entities per sentence --- higher density than religious text or children's stories --- making it a strong NER seed despite the share-alike constraint. Practitioners who need CC-BY 4.0 output should use WURA or MT560-derived corpora as seeds and accept somewhat lower entity density.

\section{Conclusion}

The four case studies share a pattern: a legal assumption was made implicitly that would not have survived explicit examination. JW300 was used because it existed and seemed open. Tanzil was treated as CC-BY because that is what the label said. WAXAL's per-provider terms were not traced because the arXiv paper did not prompt it. The CRC's YouTube dependency was not flagged because URL persistence is not a standard publication criterion. None of these errors was wilful; all were avoidable with a one-time provenance check.

Concrete outcomes from the compatibility analysis: the 27Group Feriji corpus (T4a) is incompatible with the 27Group GEC corpus (T3) despite coming from the same research group; the Leipzig \texttt{mkw\_community\_2017} entry is, to the author's knowledge, the largest open Kituba corpus and has not appeared in any published NLP work; and the Kongo Wikipedia (\texttt{kgwiki}) contains 1,200+ articles of Kituba text mislabeled as Kongo, a confusion that extends to FLORES-200's \texttt{kon\_Latn} entry. Data logging and persistent archiving should become standard publication requirements for African NLP work.

\section*{LRE Map}

This paper does not introduce new language resources; it audits the license provenance of existing ones. No new LRE Map entries are created. All resources cited here are existing catalogued resources; their existing identifiers (ISLRN, HuggingFace dataset IDs, OPUS corpus IDs, or GitHub repositories) are referenced in the bibliography. The LRE Map URL is \url{http://lremap.elra.info}.

\section*{Ethical Considerations}

This paper audits license provenance of existing resources; no new datasets, models, or personal data were collected. The legal analysis draws on published reports, license page text, and dataset cards retrieved in January--February 2026. None of the corpora identified as legally problematic (JW300, Tanzil, TED2020) were used to produce any output. Licenses may change after the retrieval date; practitioners should verify independently. This paper does not constitute legal advice. Future annotation work on the languages surveyed should follow community consent protocols as outlined by \citet{Nekoto2020Participatory}.

\section*{Limitations}

Jurisdiction-specific law (e.g., EU TDM exceptions under the DSM Directive) may alter practical conclusions; redistribution of modified corpora would remain restricted regardless. Tier assignments for MT560/HuggingFace datasets are provisional due to undocumented provenance.

\section{Bibliographical References}
\label{sec:reference}

\bibliographystyle{lrec2026-natbib}
\bibliography{rail2026-refs}

\begin{thebibliography}{1}
\expandafter\ifx\csname natexlab\endcsname\relax\def\natexlab#1{#1}\fi

\bibitem[{Wheatley et~al.(2020)}]{CRC}
Wheatley, Julian and others. 2020.
\newblock \href {https://github.com/lowerquality/crc} {\emph{Congolese Radio
  Corpus ({CRC}) for {Lingala}}}.
\newblock \textbf{Data persistence failure.} Originally claimed hundreds of
  hours of YouTube broadcast audio. Audit conducted February 2026 found 402 of
  405 YouTube IDs dead (404 errors). Reproducible content: approximately 8.3
  hours elicited LRSC speech (IPA-transcribed) + approximately 6.1 hours Radio
  Okapi broadcast audio = approximately 14.4 hours total.

\end{thebibliography}


\begin{thebibliography}{38}
\expandafter\ifx\csname natexlab\endcsname\relax\def\natexlab#1{#1}\fi

\bibitem[{Adelani et~al.(2021)Adelani, Abbott et~al.}]{Adelani2021MasakhaNER}
David~Ifeoluwa Adelani, Jade Abbott, et~al. 2021.
\newblock \href {https://aclanthology.org/2021.tacl-1.66} {{MasakhaNER}: Named
  entity recognition for {African} languages}.
\newblock In \emph{Transactions of the Association for Computational
  Linguistics}, volume~9, pages 1116--1131. MIT Press.

\bibitem[{Adelani et~al.(2022)Adelani, Carr et~al.}]{Adelani2022MasakhaNER2}
David~Ifeoluwa Adelani, Graham Carr, et~al. 2022.
\newblock \href {https://aclanthology.org/2022.emnlp-main.298} {{MasakhaNER}
  2.0: {Africa}-centric transfer learning for named entity recognition}.
\newblock In \emph{Proceedings of the 2022 Conference on Empirical Methods in
  Natural Language Processing}, pages 4488--4508. Association for Computational
  Linguistics.

\bibitem[{Agi{\'c} and Vuli{\'c}(2019)}]{Agic2019JW300}
{\v{Z}}eljko Agi{\'c} and Ivan Vuli{\'c}. 2019.
\newblock \href {https://aclanthology.org/P19-1310} {{JW300}: A wide-coverage
  parallel corpus for low-resource languages}.
\newblock In \emph{Proceedings of the 57th Annual Meeting of the Association
  for Computational Linguistics}, pages 3204--3210. Association for
  Computational Linguistics.

\bibitem[{Ba{\~n}{\'o}n et~al.(2020)Ba{\~n}{\'o}n, Chen, Haddow, Heafield,
  Hoang, Espl{\`a}-Gomis, Junczys-Dowmunt, Ma, Mathur, Paul, Roturier, and
  Sennrich}]{Banon2020ParaCrawl}
Marta Ba{\~n}{\'o}n, Pinzhen Chen, Barry Haddow, Kenneth Heafield, Hieu Hoang,
  Miquel Espl{\`a}-Gomis, Marcin Junczys-Dowmunt, Samuel Ma, Prashant Mathur,
  Paul Paul, Johann Roturier, and Rico Sennrich. 2020.
\newblock \href {https://aclanthology.org/2020.acl-main.417} {{ParaCrawl}:
  Web-scale acquisition of parallel corpora}.
\newblock In \emph{Proceedings of the 58th Annual Meeting of the Association
  for Computational Linguistics}, pages 4555--4567.

\bibitem[{Bender and Friedman(2018)}]{Bender2021DataStatementsNLP}
Emily~M Bender and Batya Friedman. 2018.
\newblock Data statements for natural language processing: Toward mitigating
  system bias and enabling better science.
\newblock \emph{Transactions of the Association for Computational Linguistics},
  6:587--604.

\bibitem[{Boyle(2003)}]{Boyle2003Enclosure}
James Boyle. 2003.
\newblock \href {https://papers.ssrn.com/sol3/papers.cfm?abstract_id=470983}
  {The second enclosure movement and the construction of the public domain}.
\newblock \emph{Law and Contemporary Problems}, 66(1/2):33--74.

\bibitem[{{Centre for Intellectual Property and Information Technology Law
  (CIPIT)}(2020)}]{CIPIT2020}
{Centre for Intellectual Property and Information Technology Law (CIPIT)}.
  2020.
\newblock \href
  {https://knowledgegov.org/masakhane-projects-use-of-the-jw300-dataset-for-natural-language-processing-copyright-issues-contract-overrides-and-cross-border-implications/}
  {Masakhane projects' use of the {JW300} dataset for natural language
  processing: Copyright issues, contract overrides and cross-border
  implications}.

\bibitem[{Diack et~al.(2026)}]{Diack2026WAXAL}
Thierno Diack et~al. 2026.
\newblock \href {https://arxiv.org/abs/2602.02734} {{WAXAL}: A large-scale
  multilingual speech dataset for {African} languages}.
\newblock \emph{arXiv preprint arXiv:2602.02734}.

\bibitem[{Dodge et~al.(2021)Dodge, Sap, Marasovic, Agnew, Ilharco, Groeneveld,
  Mitchell, and Gardner}]{Dodge2021Documenting}
Jesse Dodge, Maarten Sap, Ana Marasovic, William Agnew, Gabriel Ilharco, Dirk
  Groeneveld, Margaret Mitchell, and Matt Gardner. 2021.
\newblock \href {https://aclanthology.org/2021.emnlp-main.98} {Documenting
  large webtext corpora: A case study on the colossal clean crawled corpus}.
\newblock In \emph{Proceedings of the 2021 Conference on Empirical Methods in
  Natural Language Processing}, pages 1286--1305.

\bibitem[{Federmann et~al.(2022)Federmann, Kocmi, and Xin}]{Federmann2022NTREX}
Christian Federmann, Tom Kocmi, and Ying Xin. 2022.
\newblock \href {https://aclanthology.org/2022.sumeval-1.4} {{NTREX-128} --
  news test references for {MT} evaluation of 128 languages}.
\newblock In \emph{Proceedings of the First Workshop on Systematic Biases in
  {MT} Research}.

\bibitem[{Gebru et~al.(2021)}]{Gebru2021Datasheets}
Timnit Gebru et~al. 2021.
\newblock Datasheets for datasets.
\newblock \emph{Communications of the ACM}, 64(12):86--92.

\bibitem[{{GoingDutch.ai}(2024)}]{GoingDutch2024GEITje}
{GoingDutch.ai}. 2024.
\newblock {GEITje} takedown.
\newblock \url{https://goingdutch.ai/nl/posts/geitje-takedown/}.
\newblock Accessed February 2026.

\bibitem[{Goldhahn et~al.(2012)Goldhahn, Eckart, and
  Quasthoff}]{Goldhahn2012Leipzig}
Dirk Goldhahn, Thomas Eckart, and Uwe Quasthoff. 2012.
\newblock Building large monolingual dictionaries at the {Leipzig} corpora
  collection: From 100 to 200 languages.
\newblock In \emph{Proceedings of the Eighth International Conference on
  Language Resources and Evaluation (LREC 2012)}, pages 759--765. European
  Language Resources Association (ELRA).

\bibitem[{{Google}(2026)}]{WAXALDatasetCard}
{Google}. 2026.
\newblock {WAXAL}: A large-scale multilingual {African} language speech corpus
  -- dataset card.
\newblock \url{https://huggingface.co/datasets/google/WaxalNLP}.
\newblock Accessed February 2026. Dataset card specifies per-provider licenses:
  University of Ghana contributions are CC-BY 4.0; Makerere University, Digital
  Umuganda, Media Trust, and Loud and Clear contributions are CC-BY-SA 4.0.
  Contradicts the uniform CC-BY 4.0 claim in the arXiv paper.

\bibitem[{{GoURMET Consortium}(2020)}]{GoURMET2020}
{GoURMET Consortium}. 2020.
\newblock \href {https://gourmet-project.eu} {{GoURMET}: Generalisation of
  underrepresented languages with modern transformers and evaluation of
  robustness}.
\newblock {EU} Horizon 2020 Project 825299; lead institution: University of
  Sheffield; CC0 parallel corpora available via OPUS.

\bibitem[{Henderson et~al.(2023)Henderson, Li, Jurafsky, Hashimoto, Lemley, and
  Liang}]{Henderson2023FairUse}
Peter Henderson, Xuechen Li, Dan Jurafsky, Tatsunori Hashimoto, Mark~A. Lemley,
  and Percy Liang. 2023.
\newblock \href {https://arxiv.org/abs/2303.15715} {Foundation models and fair
  use}.
\newblock \emph{Journal of Machine Learning Research}, 24.

\bibitem[{Jernite et~al.(2022)Jernite, Nguyen, Biderman
  et~al.}]{Jernite2022Governance}
Yacine Jernite, Huu Nguyen, Stella Biderman, et~al. 2022.
\newblock \href {https://doi.org/10.1145/3531146.3534637} {Data governance in
  the age of large-scale data-driven language technology}.
\newblock In \emph{Proceedings of the 2022 {ACM} Conference on Fairness,
  Accountability, and Transparency}, pages 2206--2222.

\bibitem[{Katz(2006)}]{Katz2006Pitfalls}
Zachary Katz. 2006.
\newblock \href
  {https://ipmall.law.unh.edu/sites/default/files/hosted_resources/IDEA/idea-vol46-no3-katz.pdf}
  {Pitfalls of open licensing: An analysis of creative commons licensing}.
\newblock \emph{{IDEA}: The Intellectual Property Law Review}, 46(3).

\bibitem[{Kreutzer et~al.(2022)}]{Kreutzer2022Quality}
Julia Kreutzer et~al. 2022.
\newblock \href {https://aclanthology.org/2022.tacl-1.4} {Quality at a glance:
  An audit of web-crawled multilingual datasets}.
\newblock \emph{Transactions of the Association for Computational Linguistics},
  10:50--72.

\bibitem[{Langlais et~al.(2025)Langlais, Rosas~Hinostroza, Nee, Arnett
  et~al.}]{Langlais2025CommonCorpus}
Pierre-Carl Langlais, Carlos Rosas~Hinostroza, Mattia Nee, Catherine Arnett,
  et~al. 2025.
\newblock \href {https://arxiv.org/abs/2506.01732} {Common corpus: The largest
  collection of ethical data for {LLM} pre-training}.
\newblock \emph{arXiv preprint arXiv:2506.01732}.
\newblock Approximately two trillion tokens; multilingual, with 53\% of tokens
  from non-Western-country sources; African languages listed as a future
  expansion target (Swahili, Wolof, Bambara); web-mined components carry
  provenance questions analogous to those in WURA and CCMatrix.

\bibitem[{Lee et~al.(2023)Lee, Cooper, and Grimmelmann}]{Lee2023Talkin}
Katherine Lee, A.~Feder Cooper, and James Grimmelmann. 2023.
\newblock \href {https://arxiv.org/abs/2309.08133} {Talkin' 'bout {AI}
  generation: Copyright and the generative-{AI} supply chain}.
\newblock \emph{Journal of the Copyright Society of the {USA}}.

\bibitem[{Longpre et~al.(2023)Longpre, Mahari, Chen, Obeng-Marnu, Sileo,
  Brannon, Muennighoff, Khazam, Kabbara, Perisetla, Wu, Shippole, Bollacker,
  Wu, Villa, Pentland, and Hooker}]{Longpre2023Provenance}
Shayne Longpre, Robert Mahari, Anthony Chen, Naana Obeng-Marnu, Damien Sileo,
  William Brannon, Niklas Muennighoff, Nathan Khazam, Jad Kabbara, Kartik
  Perisetla, Xinyi Wu, Enrico Shippole, Kurt Bollacker, Tongshuang Wu, Luis
  Villa, Sandy Pentland, and Sara Hooker. 2023.
\newblock \href {https://arxiv.org/abs/2310.16787} {The data provenance
  initiative: A large scale audit of dataset licensing and attribution in
  {AI}}.
\newblock \emph{arXiv preprint arXiv:2310.16787}.

\bibitem[{Longpre et~al.(2024)Longpre, Mahari, Chen et~al.}]{Longpre2024Nature}
Shayne Longpre, Robert Mahari, Anthony Chen, et~al. 2024.
\newblock \href {https://doi.org/10.1038/s42256-024-00878-8} {The data
  provenance initiative: A large scale audit of dataset licensing and
  attribution in {AI}}.
\newblock \emph{Nature Machine Intelligence}, 6.

\bibitem[{Mahari and Longpre(2024)}]{Mahari2024Discit}
Robert Mahari and Shayne Longpre. 2024.
\newblock Discit ergo est: Training data provenance and fair use.
\newblock \emph{Network Law Review}.
\newblock Winter 2024. Also available at SSRN 4795277.

\bibitem[{Nekoto et~al.(2020)}]{Nekoto2020Participatory}
Wilhelmina Nekoto et~al. 2020.
\newblock \href {https://aclanthology.org/2020.findings-emnlp.195}
  {Participatory research for low-resourced machine translation: A case study
  in {African} languages}.
\newblock In \emph{Findings of the Association for Computational Linguistics:
  EMNLP 2020}, pages 2144--2160. Association for Computational Linguistics.

\bibitem[{{NLLB Team} et~al.(2022){NLLB Team}, Costa-juss{\`a}
  et~al.}]{NLLB2022}
{NLLB Team}, Marta~R Costa-juss{\`a}, et~al. 2022.
\newblock \href {https://arxiv.org/abs/2207.04672} {No language left behind:
  Scaling human-centered machine translation}.
\newblock \emph{arXiv preprint arXiv:2207.04672}.
\newblock {FLORES-200} benchmark included; 200 languages, CC-BY-SA 4.0.

\bibitem[{{nu.nl / Tweakers}(2024)}]{NuNL2024BREIN}
{nu.nl / Tweakers}. 2024.
\newblock Ontwikkelaar haalt {N}ederlands {AI}-taalmodel offline na verzoek
  {S}tichting {BREIN}.
\newblock
  \url{https://www.nu.nl/tweakers/6343889/ontwikkelaar-haalt-nederlands-ai-taalmodel-offline-na-verzoek-stichting-brein.html}.
\newblock Accessed February 2026.

\bibitem[{Okerie and Marivate(2024)}]{OkerieMarivate2024}
Chijioke Okerie and Vukosi Marivate. 2024.
\newblock \href
  {https://carnegieendowment.org/research/2024/04/how-african-nlp-experts-are-navigating-the-challenges-of-copyright-innovation-and-access}
  {How {African} {NLP} experts are navigating the challenges of copyright,
  innovation, and access}.
\newblock Carnegie Endowment for International Peace.

\bibitem[{Oladipo et~al.(2023)Oladipo, Idris, Anuoluwapo
  et~al.}]{Oladipo2023WURA}
Ifeoluwa~Adeyemi Oladipo, Abdulmumin Idris, Aremu Anuoluwapo, et~al. 2023.
\newblock \href {https://aclanthology.org/2023.emnlp-main.11} {Better quality
  pretraining data and {T5} models for {African} languages}.
\newblock In \emph{Proceedings of the 2023 Conference on Empirical Methods in
  Natural Language Processing}, pages 158--168. Association for Computational
  Linguistics.

\bibitem[{Omino(2025)}]{Omino2025NOODL}
Melissa Omino. 2025.
\newblock \href {https://infojustice.org/archives/46434} {The nwulite obodo
  open data license ({NOODL}): Licensing {African} datasets to support research
  and {AI} in the global south}.
\newblock Conference on Copyright and the Public Interest in Africa and the
  Global South, Johannesburg. {CIPIT}, Strathmore University.

\bibitem[{Pere{\l}kiewicz and Po{\'s}wiata(2024)}]{Perelkiewicz2024WebMined}
Micha{\l} Pere{\l}kiewicz and Rafa{\l} Po{\'s}wiata. 2024.
\newblock \href {https://arxiv.org/abs/2407.07630} {A review of the challenges
  with massive web-mined corpora used in large language models pre-training}.
\newblock \emph{arXiv preprint arXiv:2407.07630}.
\newblock ICAISC 2024. Surveys noise, duplication, bias, and legal issues in
  web-mined LLM pre-training data.

\bibitem[{Rijgersberg(2023)}]{Rijgersberg2023GEITje}
Edwin Rijgersberg. 2023.
\newblock \href {https://github.com/Rijgersberg/GEITje} {{GEITje}: A large open
  dutch language model}.
\newblock GitHub repository; model subsequently removed from HuggingFace at the
  request of Stichting BREIN due to copyright concerns over Dutch GigaCorpus
  training data.
\newblock Demonstrates that training data provenance problems are not limited
  to low-resource languages; a Dutch (high-resource) language model was taken
  down following copyright enforcement action by a national rights management
  foundation.

\bibitem[{{RTL Nieuws}(2024)}]{RTLNL2024GEITje}
{RTL Nieuws}. 2024.
\newblock Illegale dataset van zinnen uit {N}ederlandse films en boeken
  offline.
\newblock
  \url{https://www.rtl.nl/nieuws/tech/artikel/5465687/illegale-dataset-van-zinnen-uit-nederlandse-films-en-boeken-offline}.
\newblock Accessed February 2026.

\bibitem[{Schwenk et~al.(2021)Schwenk, Wenzek, Edunov, Grave, Joulin, and
  Fan}]{Schwenk2021CCMatrix}
Holger Schwenk, Guillaume Wenzek, Sergey Edunov, {\'E}douard Grave, Armand
  Joulin, and Angela Fan. 2021.
\newblock \href {https://aclanthology.org/2021.acl-long.70} {{CCMatrix}: Mining
  billions of high-quality parallel sentences on the {WEB}}.
\newblock In \emph{Proceedings of the 59th Annual Meeting of the Association
  for Computational Linguistics}, pages 932--944. Association for Computational
  Linguistics.

\bibitem[{{Tanzil Project}(2010)}]{TanzilLicense}
{Tanzil Project}. 2010.
\newblock \href {https://tanzil.net/docs/text_license} {Text license --- tanzil
  documents}.
\newblock Accessed February 2026. States: ``You are not allowed to modify this
  text in any way.''.

\bibitem[{Tiedemann(2020)}]{Tiedemann2020Tatoeba}
J{\"o}rg Tiedemann. 2020.
\newblock \href {https://aclanthology.org/2020.wmt-1.139} {The tatoeba
  translation challenge -- realistic data sets for low resource and
  multilingual {MT}}.
\newblock In \emph{Proceedings of the Fifth Conference on Machine Translation},
  pages 1174--1182.

\bibitem[{{Tweakers}(2024)}]{Tweakers2024BREIN}
{Tweakers}. 2024.
\newblock {BREIN} haalt illegale {N}ederlandstalige dataset voor trainen
  {AI}-modellen offline.
\newblock
  \url{https://tweakers.net/nieuws/225340/brein-haalt-illegale-nederlandstalige-dataset-voor-trainen-ai-modellen-offline.html}.
\newblock Accessed February 2026.

\bibitem[{{Walled Culture}(2020)}]{WalledCulture2020}
{Walled Culture}. 2020.
\newblock \href
  {https://walledculture.org/a-blatant-no-from-a-copyright-holder-stops-vital-linguistic-research-work-in-africa/}
  {A ``blatant no'' from a copyright holder stops vital linguistic research
  work in {Africa}}.

\end{thebibliography}

\section{Language Resource References}
\label{lr:ref}
\bibliographystylelanguageresource{lrec2026-natbib}
\bibliographylanguageresource{rail2026-langresources}

\appendix
\section{African NLP Corpus Survey}
\label{app:corpus_survey}

\begin{table*}[h]
\centering
\scriptsize
\setlength{\tabcolsep}{4pt}
\begin{tabular}{p{2.2cm}lp{2.8cm}p{2.4cm}p{5.2cm}}
\hline
\textbf{Corpus} & \textbf{Tier} & \textbf{License} & \textbf{African coverage} & \textbf{Notes} \\
\hline
UDHR & T1 & Public domain & 570+ languages & \textbf{Directly parallel}: same 30 articles across all editions; zero-pivot A--B pairs; available sentence-aligned via OPUS. \\
Wikidata labels & T1 & CC0 & Many languages & Structured data; not running text. \\
TICO-19 & T1 & CC0 & amh, hau, kin, lin, lug, orm, som, swc, swh, tir, zul (16 total) & 3,071 professionally translated segments; COVID domain; CC0 (not CC-BY as sometimes cited); the only OPUS resource with swc and lin as distinct codes. \\
GoURMET \citep{GoURMET2020} & T1 & CC0 & amh, hau, ibo, swa, tir, yor & EU Horizon-funded; University of Sheffield lead; institutional provenance equivalent to ParaCrawl. \\
NLLB bitext \citep{NLLB2022} & T2* & ODC-BY 1.0 & 35+ African languages incl.\ mos, lin, bam, fuv & Web-mined; ODC-BY governs database rights, not rights in the underlying text; contains Common Crawl-derived content; covers Moore and Bambara not in other T1/T2 sources. \\
FLEURS transcripts & T2 & CC-BY 4.0 & 20 sub-Saharan languages & Text transcripts of speech; high quality. \\
WURA \citep{Oladipo2023WURA} & T2* & Apache 2.0 & 16 African + eng/fra/por/arz & Built on mC4 (Common Crawl-derived) plus focused crawls; also includes English, French, Portuguese, and Arabic (Egyptian); Apache 2.0 applies to packaging only; underlying text is copyrighted web content; redistribution as published annotation seed carries higher legal risk than model training use. \\
MT560/HuggingFace & T2 & CC-BY 4.0 & Many; varies & Provenance often undocumented; treat as T2 with caution. \\
SMOL/gatitos & T2 & CC-BY 4.0 & Selected pairs incl.\ ktu & Source sentences selected from CommonCrawl; professional translations; small. \\
Common Voice & T2 & CC0 (recordings) & kin, swa, hau, yor, others & Speech only; text prompts vary. \\
Wikipedia & T3 & CC-BY-SA 4.0 & 40+ language editions & High entity density; propagates SA. \\
FLORES-200 & T3 & CC-BY-SA 4.0 & 30 African languages & 1,012 sentences/language; propagates SA. \\
Leipzig Corpora & T2* & CC-BY & 250+ languages & Downloadable sentence corpora are CC-BY; the NC restriction applies to online portal tools only; corpora are built via web crawling. \\
African Storybook & T2/T4a & CC-BY or CC-BY-NC & 60+ languages incl.\ dje & Per-story license; NC stories incompatible with T3 if mixed. \\
BibleTTS & T3 & CC-BY-SA 4.0 & aka, twi, ewe, hau, lin, yor & Speech with text alignment; SA propagates. \\
ParaCrawl & T1* & CC0 & swa (132,517), som (14,879) & EU CEF-funded; university-reviewed; CC0 covers packaging only; ParaCrawl explicitly does not own the underlying text. \\
NTREX-128 & T3 & CC-BY-SA 4.0 & 24 African languages & 1,997 professionally translated news sentences; same source enables direct African--African pairing; evaluation scale only. \\
Global Voices (OPUS) & T2 & CC-BY 3.0 & swa ($\approx$20K), amh ($\approx$1K) & Human-translated citizen journalism; no ShareAlike; only two African languages with meaningful coverage in OPUS. \\
eBible.org & T1--T3 & Variable per translation & 1,000+ languages & Must verify per translation; some T1, some T3, some T5. \\
Tanzil & T4b & CC-BY 3.0 + NoDerivs & hau, swa, som, amh, yor (partial) & ND clause explicit on licence page; mislabeled as CC-BY on aggregator sites; distributing annotated derivatives not permitted under the license. \\
27Group Feriji & T4a & CC-BY-NC 4.0 & dje (Zarma) & Incompatible with T3 (Wikipedia). \\
Mozilla TTS mkw & T4b & NOODL-1.0 & mkw (Kituba) & Restrictive; no redistribution, no AI derivatives without permission. \\
NaijaSenti & T2 & CC-BY 4.0 & hau, ibo, yor, pcm & Twitter-sourced; follow platform ToS for redistribution. \\
bible-uedin & T1 & CC0 & dje, hau, swa, amh, yor, others & Both OPUS and the source repository assert CC0. Practitioners should verify that the licensors hold the rights to apply CC0 to each translation before relying on this label. Covers Zarma (\texttt{dje}). \\
TED2020 & T4b & CC-BY-NC-ND 4.0 & swa, others & Both NC and ND; derivative annotation datasets prohibited regardless of output license. Frequently cited without noting T4b status. \\
CCMatrix & \textit{Unknown} & \textit{Unstated} & Many (automatically mined) & Mined from Common Crawl; no stated license on OPUS or in paper; derived datasets carry irresolvable provenance gap. \\
JW300 & T5 & ToS violation & 300+ languages & Prohibited; OPUS removed; provenance contamination risk. \\
\hline
\end{tabular}
\caption{License tier assignments for major African NLP corpus families. Asterisk (*) marks web-mined corpora where the dataset license covers packaging or database rights, not rights in the underlying text. Italic entries in the Tier column represent cases where the stated license differs from the effective license after review.}
\label{tab:corpus_survey}
\end{table*}

\end{document}